\documentclass[11pt]{elsarticle}

\usepackage{amssymb,amsfonts,amsmath,mathrsfs,mathtools}
\usepackage{graphicx,epsfig}
\usepackage{bm}
\usepackage[margin=1in,papersize={8.5in,11in}]{geometry}
\usepackage{times}
\usepackage{algorithm}
\usepackage{algpseudocodex}
\usepackage{caption}
\usepackage{subcaption}
\usepackage{float} 
\usepackage{adjustbox} 
\usepackage{wrapfig} 
\usepackage{layouts} 

\usepackage{tikz}
\usetikzlibrary{arrows.meta,positioning}

\title{Generative forecasting with joint probability models}

\begin{document}
\begin{frontmatter}

\author[ucsc]{Patrick Wyrod}
\ead{pwyrod@ucsc.edu}
\author[ucsc]{Ashesh Chattopadhyay}
\ead{aschatto@ucsc.edu}
\author[ucsc]{Daniele Venturi\corref{correspondingAuthor}}
\ead{venturi@ucsc.edu}

\address[ucsc]{Department of Applied Mathematics, 
University of California Santa Cruz\\ Santa Cruz (CA) 95064}

\cortext[correspondingAuthor]{Corresponding author}

\journal{arXiv}

\begin{abstract}
 Chaotic dynamical systems exhibit strong sensitivity to initial conditions and often contain unresolved multiscale processes, making deterministic forecasting fundamentally limited. Generative models offer an appealing alternative by learning distributions over plausible system evolutions; yet, most existing approaches focus on next-step conditional prediction rather than the structure of the underlying dynamics. In this work, we reframe forecasting as a fully generative problem by learning the joint probability distribution of lagged system states over short temporal windows and obtaining forecasts through marginalization. This new perspective allows the model to capture nonlinear temporal dependencies, represent multistep trajectory segments, and produce next-step predictions consistent with the learned joint distribution. We also introduce a general, model-agnostic training and inference framework for joint generative forecasting and show how it enables assessment of forecast robustness and reliability using three complementary uncertainty quantification metrics (ensemble variance, short-horizon autocorrelation, and cumulative Wasserstein drift), without access to ground truth. We evaluate the performance of the proposed method on two canonical chaotic dynamical systems, the Lorenz--63 system and the Kuramoto--Sivashinsky equation, and show that joint generative models yield improved short-term predictive skill, preserve attractor geometry, and achieve substantially more accurate long-range statistical behaviour than conventional conditional next-step models.
\end{abstract}
\end{frontmatter}


\section{Introduction}
\label{sec:intro}
Generative modelling has become a central paradigm in modern machine learning due to its ability to learn complex, high-dimensional probability distributions and generate samples representative of those distributions. Unlike discriminative models, which directly map inputs to outputs, generative models seek to approximate the underlying data-generating process itself. This capability enables sampling, uncertainty quantification, and distribution-level reasoning. They have transformed applications in computer vision, natural language processing, and scientific machine learning as a result. These strengths are particularly compelling in settings where the underlying processes are stochastic, only partially observed, or governed by incomplete physical laws---contexts in which a single deterministic prediction cannot adequately represent the evolution of the system. The central motivation behind generative forecasting is the recognition that \emph{model insufficiency/inadequacy is unavoidable} in high-dimensional chaotic dynamical systems without scale separation. In fact, unresolved subgrid processes, imperfect parameterizations, truncated modes, and limited training data all create epistemic uncertainty that accumulates and amplifies unpredictability under chaotic dynamics. Traditionally, generative models address this by learning a distribution over physically plausible trajectories conditioned on known information. In this sense, probabilistic forecasting becomes a principled framework for representing the ensemble of futures consistent with the available data and with the intrinsic variability of the system.

Forecasting chaotic dynamical systems exemplifies this setting. High-dimensional, multiscale chaotic systems arising in climate, atmosphere, ocean, and turbulence are extremely sensitive to initial conditions and contain unresolved physical processes that no deterministic model, numerical or neural, can perfectly capture \cite{Ashesh1,Ashesh2}. These structural deficiencies motivate a shift from point forecasting to \emph{probabilistic forecasting}, where the objective is to characterize a distribution over plausible futures rather than a single trajectory. Recent research highlights the suitability of generative models for this task. For example, Chattopadhyay et al.~\cite{chattopadhyay2023long} demonstrated that generative surrogates can produce statistically consistent long-term trajectories for canonical multiscale chaotic systems. The GenCast model~\cite{gencast} shows that fully data-driven conditional diffusion models can generate accurate ensemble weather forecasts by learning the distribution of next-step atmospheric states conditioned on the previous ones. DYffusion~\cite{dyffusion} proposes generative forecasting through a predictor--corrector scheme leveraging a pair of neural network estimators to forecast several steps into the future and interpolate for the intermediate steps. While these are novel approaches to forecasting directly employing generative models, they are built upon \emph{conditional} generative models directly modelling the target density. Other hybrid approaches explicitly combine deterministic forecasting with generative modelling: G-LED~\cite{gled} uses a generative diffusion model to represent unresolved small-scale dynamics, while the ``thermalizer'' approach in Pedersen et al.~\cite{pedersen2025thermalizer} stabilizes deterministic forecasts by nudging unstable states back toward plausible initial conditions sampled from a trained generative model. Collectively, these lines of work underscore a key insight: generative stochasticity is not merely noise, but an essential mechanism for representing uncertainty and missing physics in chaotic systems.

In this work, we extend this philosophy by \emph{modelling the dynamics as a generative process} similar to the approaches adopted in Sambamurthy {\em et al.}~\cite{sambamurthy2025lazy} and Ruhling {\em et al.}~\cite{dyffusion,ruhling2024probablistic}. 
However, a key difference in our approach is that instead of directly learning a conditional next-step map $p(x_{t} \mid x_{t-\Delta t}, \ldots)$, we learn the {\em joint distribution} over multiple adjacent time steps and treat forecasting as a marginalization problem. This reframes forecasting as a \emph{purely generative task}: the next-step prediction corresponds to selecting (via marginalization) the appropriate conditional component of a learned joint distribution. Modelling short temporal windows jointly allows the generative model to capture nonlinear dependencies and correlations between neighbouring states, enabling richer representations of uncertainty \cite{DIAMZON2026108178} and more robust sampling.
A major contribution of this paper is the use of the joint probability distributions to perform intrinsic uncertainty quantification. Because each generative sample corresponds to a plausible short trajectory segment, the geometry of the sampled point cloud encodes valuable information about the model’s confidence. Leveraging this structure, we introduce several joint probability-driven uncertainty metrics, including ensemble variance, short-horizon autocorrelation, and a cumulative Wasserstein drift. Collectively, these allow us to estimate forecast reliability \emph{a priori}, without requiring ground truth observations or running additional inference passes. This capability is of particular importance for long-range forecasting, model diagnostics, and out-of-distribution stability assessment.

Beyond uncertainty quantification, we evaluate forecasting performance through traditional metrics such as short-term skill, long-term statistical consistency, and the ability to extrapolate to extremes beyond the training distribution. The latter is especially crucial for long-term emulation of the coupled climate system, as emphasized in Sun et al.~\cite{sun2025can}, where tail behaviour and extreme events play a central scientific role. Our generative joint modelling framework unifies these goals, enabling robust prediction, principled uncertainty representation, and diagnostic interpretability within a single modelling paradigm.
In summary, our contributions are threefold: (1) a fully generative forecasting framework based on modelling joint temporal distributions and extracting next-step predictions via marginalization, (2) general and model-agnostic training and inference procedures compatible with any generative model architecture, and (3) novel uncertainty metrics derived from joint probabilities which can diagnose forecast quality without observational data. Together, these contributions demonstrate that forecasting can be formulated fundamentally as a generative problem, yielding improved uncertainty representation, stronger long-term stability, and deeper insights into the behaviour of data-driven dynamical system models.

This paper is organized as follows: In Section~\ref{sec:methods}, we describe the proposed joint generative forecasting framework, including the training procedure, inference via joint marginalization, and uncertainty quantification based on joint samples. In Section~\ref{sec:results}, we demonstrate the performance of the generative forecasting algorithm using synthetic data from two canonical chaotic dynamical systems, namely the Lorenz--63 system and the Kuramoto--Sivashinsky equation. The main findings and their implications are summarized in Section~\ref{sec:conclusion}.

\begin{figure}[t]
  \centering
  \input{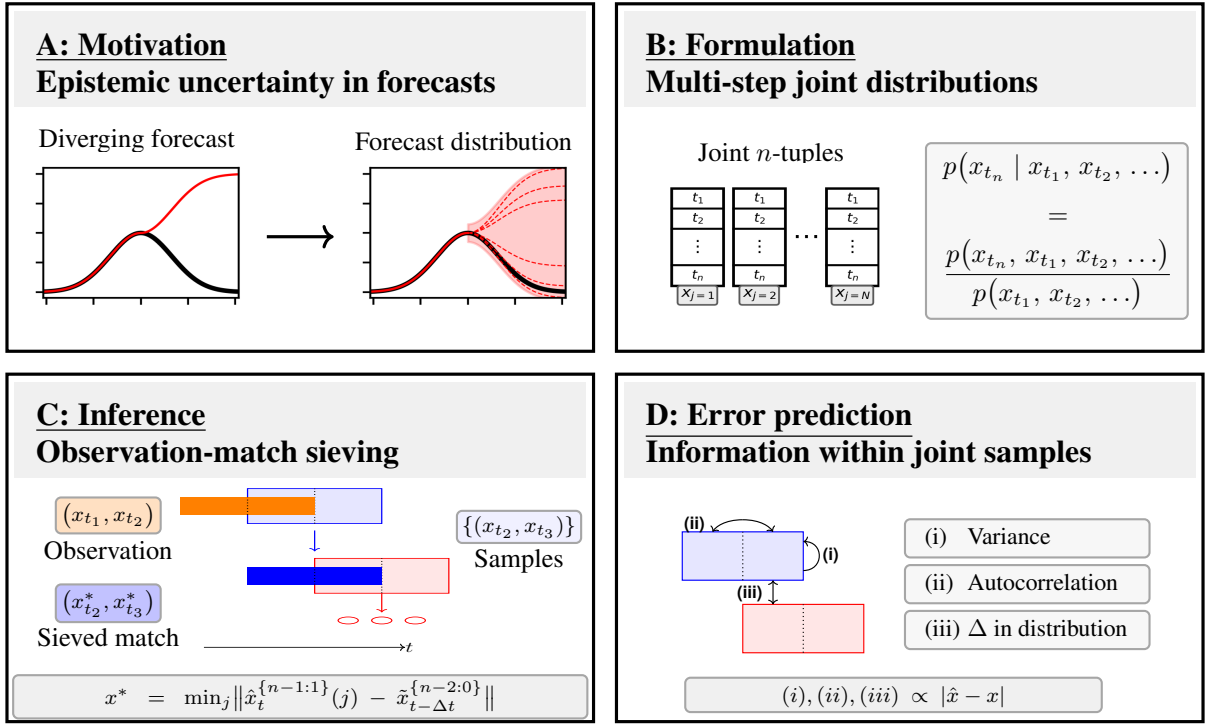}
 \caption{Schematic overview of the proposed generative joint forecasting framework for chaotic dynamical systems. (A) Inherent challenges of forecasting high-dimensional chaotic systems, characterized by extreme sensitivity to initial conditions and complex multi-scale dynamics. (B) Proposed core methodology: modelling the joint probability distribution of temporal sequences, e.g., $p({x}_{t_n}, {x}_{t_1}, {x}_{t_2},\ldots)$. The forecast for the current state, ${x}_{t_n}$, is then obtained through marginalization of this learned joint distribution. (C) Forecasts are obtained by sieving a joint ensemble using components overlapping with an observed segment. (D) Joint samples enable intrinsic uncertainty quantification through ensemble variance, autocorrelation, and Wasserstein drift metrics.}
  \label{fig:viz_abstract}
\end{figure}

\section{Methods}
\label{sec:methods}
In Figure \ref{fig:viz_abstract}, we provide a high-level overview of the proposed generative forecasting method based on joint probability models.
Forecasting can be viewed as predictive statistical inference on ordered data. Given a time series $\{x_t\}_{t}$ with temporal spacing $\Delta t$, a forecasting model ultimately targets the conditional probability density 
\begin{equation}
    \label{eq:forecasting_obj}
    \hat{p}_\theta(x_t)
    \;\approx\;
    p\bigl(x_t \mid x_{t-\Delta t} = \tilde{x}_{t-\Delta t},\, x_{t-2\Delta t} = \tilde{x}_{t-2\Delta t},\, \ldots\bigr),
\end{equation}
where $\tilde{x}_{t-k\Delta t}$ denotes the observed state at time $t-k\Delta t$.
Traditional approaches seek to directly parameterize this conditional distribution. However,  $p\bigl(x_t \mid x_{t-\Delta t} = \tilde{x}_{t-\Delta t},\, x_{t-2\Delta t} = \tilde{x}_{t-2\Delta t},\, \ldots\bigr)$ can equivalently be written in terms of joint distributions over a short temporal window as 
\begin{equation}
    \label{eq:full_joint_obj}
    p\bigl(x_t \mid x_{t-\Delta t}=\tilde{x}_{t-\Delta t},\,x_{t-2\Delta t}=\tilde{x}_{t-2\Delta t},\,\ldots\bigr)
    =
    \frac{
      p\bigl(x_t,\,x_{t-\Delta t}=\tilde{x}_{t-\Delta t},\,x_{t-2\Delta t}=\tilde{x}_{t-2\Delta t},\,\ldots\bigr)
    }{
      p\bigl(x_{t-\Delta t}=\tilde{x}_{t-\Delta t},\,x_{t-2\Delta t}=\tilde{x}_{t-2\Delta t},\,\ldots\bigr)
    }.
\end{equation}
Generative models can be naturally defined in terms of such joint distributions. We therefore propose to model the joint distribution over a short sequence of adjacent time steps,
\begin{equation}
    \label{eq:joint_obj}
    \hat{p}_\theta\left(x_t,\,x_{t-\Delta t},\,x_{t-2\Delta t},\,\ldots\right)
    \approx
    p\left(x_t,\,x_{t-\Delta t},\,x_{t-2\Delta t},\,\ldots\right),
\end{equation}
and to obtain the forecasting objective, i.e.\ the conditional density given by Eq.~\eqref{eq:forecasting_obj}, at inference time by marginalization and conditioning via Eq.~\eqref{eq:full_joint_obj}.
%

Within this framework, $\hat{p}_\theta$ can be realized using any suitable generative modelling paradigm (e.g.\ variational autoencoders~\cite{kingma2022autoencodingvariationalbayes}) in conjunction with any estimator backbone (such as simple feedforward neural networks or transformers~\cite{vaswani2023attentionneed}). The generative model learns to sample from $\hat{p}_\theta$, rather than to approximate its functional form explicitly.
The key structural feature is that the state dimension $d$ and the number of jointly modelled time steps $n$ reside on distinct axes and can therefore be handled separately by the architecture. For low-dimensional or non-complex data, a simple concatenation-based estimator architecture may suffice; for high-dimensional or long-horizon data, sequence-aware estimators are preferable.

\subsection{Training protocol}
\label{subsec:jgf_training}

The temporal window length $n$ is a hyperparameter of the method. Larger $n$ allows the model to exploit longer temporal correlations but increases the difficulty of the learning problem and hence the required model capacity. Given $n$ and a time step $\Delta t$, we construct training samples as short sequences of adjacent states.
One must select both a generative modelling technique and a neural network estimator backbone well-suited to model~\eqref{eq:joint_obj} for a given application. For sequence-aware models, such as those built upon recurrent neural networks or transformers, the input at time $t$ is the ordered sequence defined as
\begin{equation}
    \label{eq:seq_input}
    x_t^{\{n-1:0\}}
    =
    \bigl(x_{t-(n-1)\Delta t},\,\ldots,\,x_{t-\Delta t},\,x_t\bigr)
    \in \mathbb{R}^{n \times d},
\end{equation}
while for models that operate on a single vector, the same information is provided via concatenation:
\begin{equation}
    \label{eq:seq_input_concat}
    x_t^{\{n-1:0\}}
    =
    \bigl[x_{t-(n-1)\Delta t},\,\ldots,\,x_{t-\Delta t},\,x_t\bigr]
    \in \mathbb{R}^{nd}.
\end{equation}
In either case, the generative model $\mathcal{J}_\theta$ is trained to sample from the joint distribution  \eqref{eq:joint_obj} over $n$ adjacent states,
so that drawing a sample from the model yields
\begin{equation}
    x_t^{\{n-1:0\}} = \mathcal{J}_\theta(z), \qquad   x_t^{\{n-1:0\}}\sim 
    \hat{p}_\theta\left(x_t,\,x_{t-\Delta t},\,x_{t-2\Delta t},\,\ldots\right),
\end{equation}
where the input $z$ is a sample from the generative model's latent space (see also Figure~\ref{fig:vae_schematic} for an example).
Throughout this work, we focus on uniformly spaced temporal data with fixed $\Delta t$, but the formulation easily generalizes by including the time step size as an explicit conditioning variable,
\begin{equation}
    \label{eq:joint_n_nonuniform}
    \hat{p}_\theta
    \;\approx\;
    p\left(x_t,\,x_{t-\Delta t},\,\ldots,\,x_{t-(n-1)\Delta t},\,\Delta t\right),
\end{equation}
and, if desired, additional parameters for nonautonomous systems.

\subsection{Inference via joint marginalization}
\label{subsec:jgf_inference}

\begin{figure}[t]
  \centering
  \includegraphics[width=0.8\linewidth]{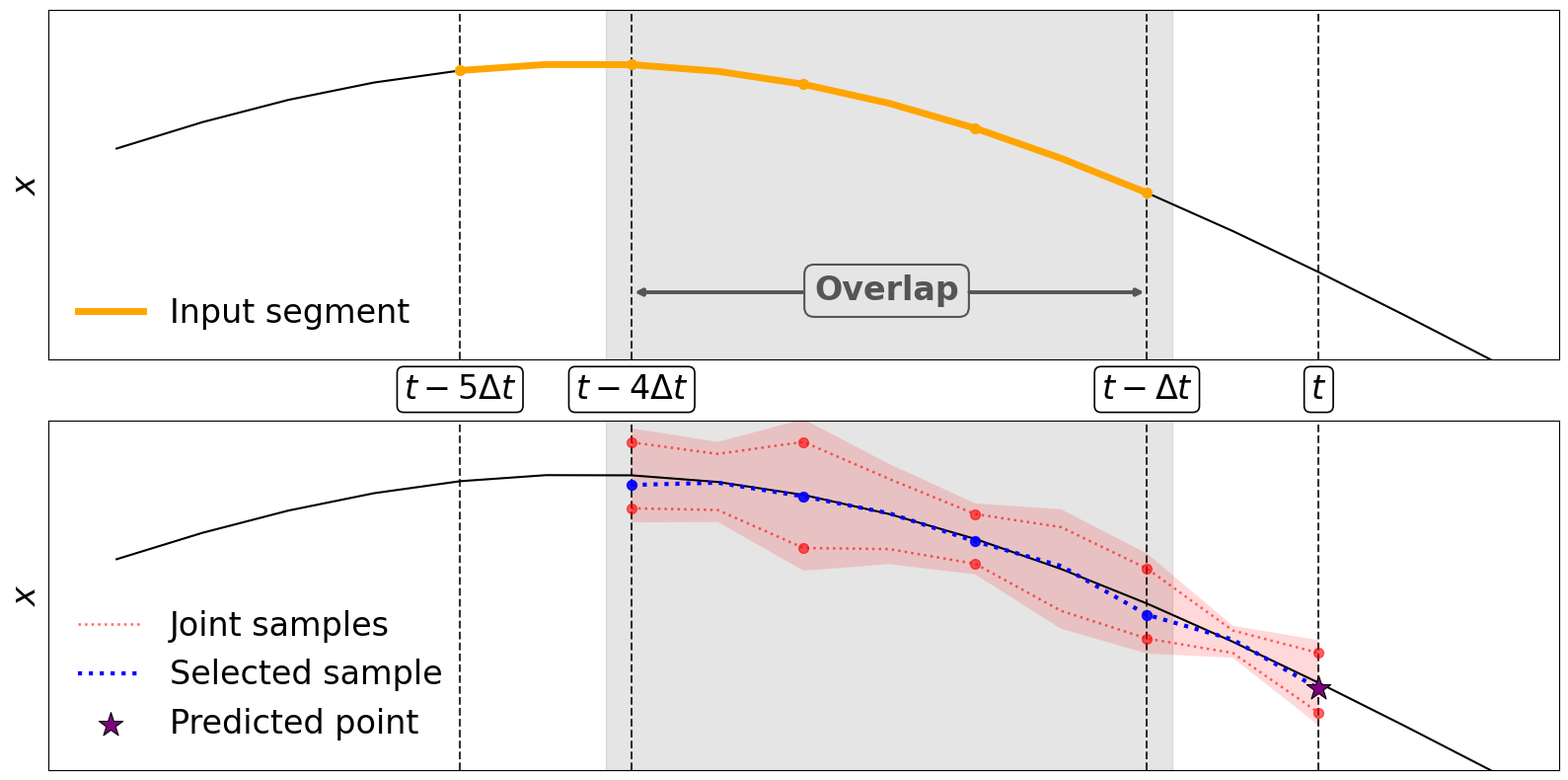}
  \caption{Visualization of a single inference step with Algorithm~\ref{alg:inference} for $n=5$. The trajectory formed by the orange points (top) serves as the reference for isolating the closest-matching tail points, $\hat{x}_{t-(n-1)\Delta t}$ through $\hat{x}_{t-\Delta t}$, from the joint point cloud (bottom). The corresponding head point $\hat{x}_t$ (purple star) is the prediction, which can then be appended to the input set (orange) to perform a successive autoregressive step.}
  \label{fig:alg1_visual}
\end{figure}

Once the joint distribution, given by Eq.~\eqref{eq:joint_obj}, has been modelled, we obtain point forecasts by marginalizing an ensemble of joint samples. The basic procedure is summarized in Algorithm~\ref{alg:inference} and illustrated in Figure~\ref{fig:alg1_visual}. At each autoregressive step, we draw a point cloud of $N$ joint samples from the unconditional model and select the sample whose ``tail'' subsequence best matches the most recent observed history on a chosen metric (e.g. Euclidean distance); the corresponding ``head'' state is used as the forecast.
\begin{algorithm}[t]
    \caption{Forecasting through marginalization}
    \label{alg:inference}
    \begin{algorithmic}[1]
      \State \textbf{Input:} $\mathcal{J}_\theta,\ \tilde{x}_{t-\Delta t}^{\{n-1:0\}}$ \Comment{trained model \& observed $n$-lag sequence}
      \State \textbf{Initialization:} Sample $\left\{\hat{x}_t^{\{n-1:0\}}(j)\right\}_{j=1}^N \sim \mathcal{J}_\theta$
      \Repeat
        \State $j^* \gets \displaystyle {\arg\min}_{j} \left\|\hat{x}_{t}^{\{n-1:1\}}(j)-\tilde{x}_{t-\Delta t}^{\{n-2:0\}}\right\|$ \Comment{match on trailing $n-1$ states}
        \State retain $x_t^* \gets \hat{x}_t^{\{0\}}(j^*)$
        \State update observation $\tilde{x}_t^{\{n-1:0\}} \gets \tilde{x}_{t-\Delta t}^{\{n-1:1\}} \oplus x_t^*$ \Comment{shift window and append forecast}
        \State $t \gets t + \Delta t$
      \Until{desired forecast horizon reached}
      \State \textbf{Return:} retained $\{x_t^*\}$
    \end{algorithmic}
\end{algorithm}

Each sample corresponds to a short, self-consistent trajectory segment of length $n$, and the argmin step selects the segment whose past matches the current observed history. Formally, we draw
\begin{equation}
    \label{eq:j_theta_cloud}
    \left\{\left(\hat{x}_t,\,\hat{x}_{t-\Delta t},\,\ldots,\,\hat{x}_{t-(n-1)\Delta t}\right)(j)\right\}_{i=1}^N
    \sim \hat{p}_\theta\left(x_t,\,x_{t-\Delta t},\,\ldots,\,x_{t-(n-1)\Delta t}\right),
\end{equation}
then isolate
\[
    \left(x_t^*,\,x_{t-\Delta t}^*,\,\ldots,\,x_{t-(n-1)\Delta t}^*\right) = \left(\hat{x}_t,\,\hat{x}_{t-\Delta t},\,\ldots,\,\hat{x}_{t-(n-1)\Delta t}\right)(j^*)
\]
using overlapping entries with the current observed sequence of states
\[
    j^* = {\arg\min}_{j} \left\|\hat{x}_{t}^{\{n-1:1\}}(j)-\tilde{x}_{t-\Delta t}^{\{n-2:0\}}\right\|,
\]
and use the leading entry to realize the conditional forecast
\[
    x_t^* \;\sim\; p\left(x_t \mid x_{t-\Delta t}=\tilde{x}_{t-\Delta t},\,x_{t-2\Delta t}=\tilde{x}_{t-2\Delta t},\,\ldots,\,x_{t-(n-2)\Delta t}=\tilde{x}_{t-(n-2)\Delta t}\right).
\]
This inference scheme uses an \emph{unconditional} generative model: because all samples are i.i.d.\ draws from the same joint distribution, the point cloud $\{\hat{x}_t^{\{n-1:0\}}(j)\}_{j=1}^N$ can be sampled once and reused across all autoregressive steps, or resampled at each autoregressive step, trading computation for memory as needed.

\subsection{Uncertainty quantification from joint samples}
\label{subsec:jgf_uq}

Beyond point prediction, the joint point cloud produced by $\mathcal{J}_\theta$ encodes rich information about short-horizon uncertainty and model self-consistency. To exploit this, we retain not only the best-matching sample but also its $k$ nearest neighbours in the matching metric. Concretely, at each step we define the subset
\begin{equation}
    \label{eq:subcloud}
    \left\{\left(x_t^*,\,x_{t-\Delta t}^*,\,\ldots,\,x_{t-(n-1)\Delta t}^*\right)(i)\right\}_{i=1}^k
    =
    \text{top-}k\ \left[{\arg\min}_{j} \left\|\hat{x}_{t}^{\{n-1:1\}}(j)-\tilde{x}_{t-\Delta t}^{\{n-2:0\}}\right\|\right],
\end{equation}
where \(\hat{x}_{t}^{\{n-1:1\}}(j)\) are obtained from~\eqref{eq:j_theta_cloud} and \(k \leq N\).
The first element ($i=1$) is used for the point forecast, while all $k$ elements taken together form a local ensemble drawn from intermediate posteriors
\begin{equation*}
    \left\{\hat{p}_\theta\left(x_t,\, x_{t-\Delta t}=\tilde{x}_{t-\Delta t}+\epsilon^{(j)}_{t-\Delta t},\,\ldots,\,x_{t-(n-2)\Delta t}=\tilde{x}_{t-(n-2)\Delta t}+\epsilon^{(j)}_{t-(n-2)\Delta t}\right)\right\}_{j=1}^k.
\end{equation*}
Here, \(\left\{\epsilon^{(j)}\right\}_{j=1}^k\) quantify the mismatches between the selected ensemble members in~\eqref{eq:subcloud} and the observed \(\tilde{x}\).
From this ensemble we construct the following three complementary uncertainty metrics.

\paragraph{Ensemble variance}
The spread of the ensemble around its mean provides a natural measure of uncertainty. Let $x_t(i)$ denote the $i$th ensemble member at time $t$, and let $w_i$ be optional importance weights. We define
\begin{equation}
    \label{eq:ensvar_def}
    \sigma_\text{ens}(x_t)
    =
    \frac{\displaystyle \sum_{i=1}^k
      w_i\,\bigl(x_t(i)-\bar{x}_t\bigr)^{\odot 2} }{\displaystyle\sum_{i=1}^k w_i},
    \qquad
    \bar{x}_t
    =
    \frac{\displaystyle \sum_{i=1}^k
      w_i\,x_t(i)}{\displaystyle \sum_{i=1}^k w_i},
\end{equation}
where $\odot$ denotes elementwise multiplication (Hadamard product). Unlike the subsequent metrics, ensemble variance is not specific to joint modelling; it is a generic property of any generative forecaster. However, the ranking induced by the matching step to create the ensemble as specific in Eq.~\eqref{eq:subcloud} allows us to compute $\sigma_\text{ens}$ on a continuum of posteriors, from tightly conditioned to loosely conditioned ensembles.

\paragraph{Autocorrelation}
Each joint sample in the ensemble provides a pairwise correspondence between $x_t$ and $x_{t-\Delta t}$. This allows us to estimate the linear dependence between successive states via an empirical autocorrelation:
\begin{equation}
    \label{eq:ac_def}
    AC(x_t, x_{t-\Delta t})
    =
    \frac{
    \displaystyle \frac{1}{k}
      \sum_{i=1}^k
        \bigl(x_t^{(i)}-\bar{x}_t\bigr)\odot
        \bigl(x_{t-\Delta t}^{(i)}-\bar{x}_{t-\Delta t}\bigr)
    }{
      \sigma_t \odot \sigma_{t-\Delta t}
    },
\end{equation}
where $\bar{x}_t, \bar{x}_{t-\Delta t}$ and $\sigma_t, \sigma_{t-\Delta t}$ are the empirical means and standard deviations of the ensemble at times $t$ and $t-\Delta t$, respectively. Although other dependence measures (e.g.\ mutual information) could be used, $AC$ is attractive for its interpretability and low computational cost.

\paragraph{Wasserstein drift}
In addition to pairwise structure, the joint model provides redundant marginal information across successive time steps: for a given physical time $t$, the ``head'' component of one joint sample and the ``tail'' component of the next joint sample both correspond to the same state $x_t$.
We quantify the self-consistency of these overlapping marginals using a Wasserstein distance between their empirical distributions. Let $\hat{p}_t$ and $\hat{p}_{t-\Delta t}$ denote the empirical marginals at time $t$ formed from the relevant heads and tails. We define
\begin{equation}
    \label{eq:wd_def}
    WD_t = W_2\left(\hat{p}_t,\hat{p}_{t-\Delta t}\right),
\end{equation}
where $W_2$ is the 2-Wasserstein distance, estimated in practice using a Sinkhorn-regularized optimal transport solver~\cite{chizat2020fasterwassersteindistanceestimation}. An increasing $WD_t$ indicates that the modelled marginals corresponding to the same physical time are drifting apart, suggesting growing uncertainty or inconsistency.

To relate this drift to forecast error, we construct a signed version of the Wasserstein distance using changes in ensemble variance:
\begin{equation}
    \label{eq:signed_wasserstein}
    \widetilde{WD}_t
      =
      \begin{cases}
           -\,WD_t, & \sigma_\text{ens}(x_{t+\Delta t}) < \sigma_\text{ens}(x_t),\\[4pt]
           \phantom{-}WD_t, & \text{otherwise},
      \end{cases}
\end{equation}
and define the cumulative reconstruction
\begin{equation}
    \label{eq:wdrecon_def}
    \mathrm{WD}_{\mathrm{recon}}(x_t;\{x_s : s \leq t\})
      =
      \sum_{s=1}^{t}\widetilde{WD}_s.
\end{equation}
This scalar time series serves as a correlate of the accumulated forecast error and is used in Section~\ref{sec:results} to assess the ability of our uncertainty metrics to predict pointwise mean absolute error (MAE) a priori.

\subsection{Conditional joint models and latent optimal control}
\label{subsec:jgf_extensions}

The joint generative framework is flexible and admits several useful extensions. Here we highlight two: conditional joint modelling and latent optimal control.

\paragraph{Conditional joint models}
While our baseline formulation is fully generative and unconditional, the joint target~\eqref{eq:joint_obj} can also be formulated as a \emph{conditional joint} distribution,
\begin{equation}
    \label{eq:cond_joint_obj}
    \hat{p}_\theta\bigl(x_t,\, x_{t-\Delta t},\, \ldots,\, x_{t-(n-1)\Delta t}\bigr)
    \;\approx\;
    p\left(x_t,\, x_{t-\Delta t},\, \ldots,\, x_{t-(n-1)\Delta t} \mid x_{t-\Delta t},\, x_{t-2\Delta t},\, \ldots,\, x_{t-(n-1)\Delta t}\right),
\end{equation}
and used as a drop-in replacement in the marginalization step~\eqref{eq:full_joint_obj}. In this case, the observed history appears both as an explicit conditioning input and implicitly through the marginalization of the joint distribution. By contrast, the unconditional formulation conditions solely through the selection step in Algorithm~\ref{alg:inference}. We compare unconditional and conditional variants empirically in Section~\ref{sec:results}.

\paragraph{Latent optimal control}
For high-dimensional states or long sequences, Algorithm~\ref{alg:inference} may suffer from the curse of dimensionality: an impractically large point cloud may be required to ensure a close enough match in the argmin step. When the underlying generative model admits a differentiable sampling map $\mathcal{J}_\theta(z)$ from a latent variable $z$, we can refine the inference procedure by replacing discrete sieving with optimization in latent space. The resulting \emph{latent optimal control} strategy is summarized in Algorithm~\ref{alg:zopt}.
\begin{algorithm}[t]
    \caption{Forecasting via latent optimal control}
    \label{alg:zopt}
    \begin{algorithmic}[1]
        \State \textbf{Input:} Observed $\tilde{x}_t^{\{n-1:0\}}$, initial latent $\tilde{z}_t$ such that $\mathcal{J}_\theta(\tilde{z}_t)=\tilde{x}_t^{\{n-1:0\}}$ 
        \State \textbf{Optimize:} update $z$ by gradient descent on
        \[
          \mathcal{L}(z)
          =
          \left\|\left(\mathcal{J}_\theta(z)\right)^{\{n-1:1\}}-\tilde{x}_t^{\{n-1:1\}}\right\|,
        \]
        starting from $z \gets \tilde{z}_t$, to obtain $z_t^{*}$
        \State \textbf{Return:} $x_t^{*} = \left(\mathcal{J}_\theta(z_t^{*})\right)^{\{0\}}$
    \end{algorithmic}
\end{algorithm}
Latent optimal control requires only a single sample (effectively $N=1$) and can be combined with Algorithm~\ref{alg:inference} by using a sieved sample as the initial latent. This hybrid approach retains the interpretability of the joint point cloud perspective while mitigating sparsity issues in very high-dimensional settings.

\begin{figure}[t]
\centering
\includegraphics[width=0.8\linewidth]{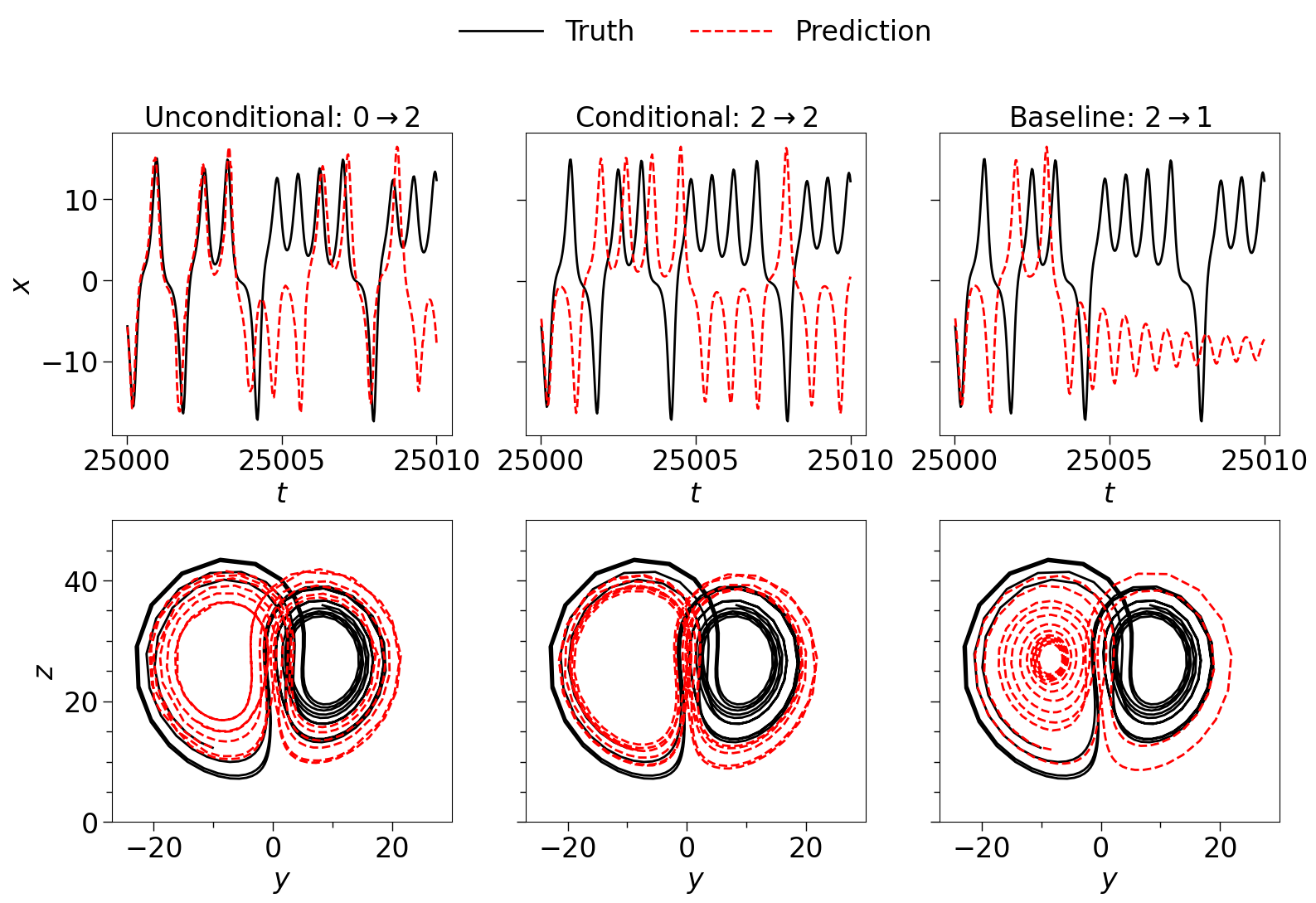}
\caption{
Single trajectory of Lorenz--63, 10 time units forward from \(t=2.5\times10^4\). Top row shows each models' \(x\) in time series, with the corresponding \((y, z)\)-phase space view below.}
  \label{fig:l63_single}
\end{figure}

\section{Numerical results}
\label{sec:results}

We evaluate the proposed joint generative forecasting framework on synthetic data from two canonical chaotic dynamical systems: Lorenz--63~\cite{Lorenz1963} and the Kuramoto--Sivashinsky (KS) equation~\cite{10.1143/PTPS.64.346,1977AcAau...4.1177S}. Chaotic systems are a natural testbed for generative forecasting: they exhibit strong sensitivity to initial conditions, multi-scale structure, and accumulation of epistemic uncertainty, while still allowing access to a well-defined ground truth trajectory for quantitative assessment. Although we focus on chaotic dynamics as a principal application setting, the method is fully data-driven and applies to general time series forecasting problems.

Throughout this section we compare three forecasting configurations:
\begin{enumerate}
    \item \textbf{Unconditional joint \,(0\,$\rightarrow$\,2)}:\quad $\hat{p}_\theta(x_t,\,x_{t-\Delta t})$,
    \item \textbf{Conditional joint\quad\, (2\,$\rightarrow$\,2)}:\quad $\hat{p}_\theta(x_t,\,x_{t-\Delta t} \mid x_{t-\Delta t},\, x_{t-2\Delta t})$,
    \item \textbf{Baseline conditional (2\,$\rightarrow$\,1)}:\quad $\hat{p}_\theta(x_t \mid x_{t-\Delta t},\, x_{t-2\Delta t})$,
\end{enumerate}
where \((a\rightarrow b)\) denotes an input sequence of length \(a\) and output of length \(b\).
The first two are instances of the joint generative framework (unconditional and conditional variants), while the third is a conventional conditional next-step model; all implementations take the simplest case of \(n=2\). Inference for the unconditional model in the short-term setting is performed via Algorithm~\ref{alg:zopt}. All other configurations and experiments use Algorithm~\ref{alg:inference} with an ensemble size of \(5\times10^4\).
We report both short-term deterministic results and long-term statistics, followed by an analysis of uncertainty-aware error prediction based on the joint samples.

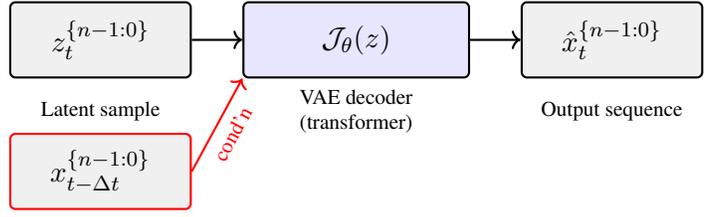
\begin{figure}[t]
  \centering
\begin{tikzpicture}[x=1.0cm,y=1.0cm,font=\sffamily]
\path[use as bounding box] (-0.2,-5.7) rectangle (16.2,2.7);

\tikzset{
box/.style={draw=black,thick,rounded corners=2pt,fill=black!6,minimum width=2.4cm,minimum height=1.0cm,align=center},
boxr/.style={draw=red,thick,rounded corners=2pt,fill=black!6,minimum width=2.4cm,minimum height=1.0cm,align=center},
nn/.style={draw=black,thick,rounded corners=2pt,fill=blue!10,minimum width=3.0cm,minimum height=1.0cm,align=center},
lbl/.style={font=\scriptsize,align=center},
hdr/.style={font=\bfseries\small,anchor=west},
cond/.style={text=red,font=\scriptsize}
}

\def\xin{1.6}
\def\xenc{5.0}
\def\xz{8.4}
\def\xdec{11.8}
\def\xout{15.2}

\def\ytrainhdr{2.25}
\def\ytrain{1.1}
\def\ytrainlbl{0.15}
\def\ytrainextra{-0.85}

\def\yinfhdr{-2.05}
\def\yinf{-3.30}
\def\yinflbl{-4.25}
\def\yinfextra{-5.05}

\node[hdr] at (0.25,\ytrainhdr) {\underline{Training}};

\node[box] (xin) at (\xin,\ytrain) {$x_t^{\{n-1:0\}}$};
\node[nn] (enc) at (\xenc,\ytrain) {$q_\phi(x)$};
\node[box] (z) at (\xz,\ytrain) {$z_t^{\{n-1:0\}}$};
\node[nn] (dec) at (\xdec,\ytrain) {$\mathcal{J}_\theta(z)$};
\node[box] (xout) at (\xout,\ytrain) {$\hat{x}_t^{\{n-1:0\}}$};

\draw[->,thick,black] (xin.east) -- (enc.west);
\draw[->,thick,black] (enc.east) -- (z.west);
\draw[->,thick,black] (z.east) -- (dec.west);
\draw[->,thick,black] (dec.east) -- (xout.west);

\node[lbl] at (\xin,\ytrainlbl) {Input sequence~\eqref{eq:seq_input}};
\node[lbl] at (\xenc,\ytrainlbl) {VAE encoder\\(transformer)};
\node[lbl] at (\xz,\ytrainlbl) {Latent mapping};
\node[lbl] at (\xdec,\ytrainlbl) {VAE decoder\\(transformer)};
\node[lbl] at (\xout,\ytrainlbl) {Output sequence};

\node[boxr] (xshift) at (\xin,\ytrainextra) {$x_{t-\Delta t}^{\{n-1:0\}}$};
\node[boxr] (zshift) at (\xz,\ytrainextra) {$x_{t-\Delta t}^{\{n-1:0\}}$};
\draw[->,thick,red] (xshift.east) -- (enc.south west) node[midway,below,sloped,cond]{cond'n};
\draw[->,thick,red] (zshift.east) -- (dec.south west) node[midway,below,sloped,cond]{cond'n};

\draw[black!55,thick] (0.25,{\yinfhdr+0.35}) -- (15.95,{\yinfhdr+0.35});

\node[hdr] at (0.25,\yinfhdr) {\underline{Inference}};

\node[box] (zsample) at (\xz,\yinf) {$z_t^{\{n-1:0\}}$};
\node[nn] (dec2) at (\xdec,\yinf) {$\mathcal{J}_\theta(z)$};
\node[box] (xout2) at (\xout,\yinf) {$\hat{x}_t^{\{n-1:0\}}$};

\draw[->,thick,black] (zsample.east) -- (dec2.west);
\draw[->,thick,black] (dec2.east) -- (xout2.west);

\node[lbl] at (\xz,\yinflbl) {Latent sample};
\node[lbl] at (\xdec,\yinflbl) {VAE decoder\\(transformer)};
\node[lbl] at (\xout,\yinflbl) {Output sequence};

\node[boxr] (zshift2) at (\xz,\yinfextra) {$x_{t-\Delta t}^{\{n-1:0\}}$};
\draw[->,thick,red] (zshift2.east) -- (dec2.south west) node[midway,below,sloped,cond]{cond'n};

\end{tikzpicture}
  \caption{Schematic overview of the specific model implementation throughout Section~\ref{sec:results}. The transformers (blue) are configured to have an input dimension of 256 (8 reserved for sinusoidal position encoding), feedforward dimension of 1024, 4 layers, and 4 attention heads. They are configured as encoder--decoder architectures for the conditional models and decoder-only for the unconditional model; the input and latent sequences are inputs for their respective transformers' decoders. Red outlines denote the conditioning inputs passed into the encoder portion of the transformers (when present).}
  \label{fig:vae_schematic}
\end{figure}

A common transformer-based variational autoencoder (VAE) \(\mathcal{J}_\theta\) backbone is held constant across all model and experiment configurations.
In particular, \(\mathcal{J}_\theta\) is selected such that conditioning information is ingested through a transformer encoder, while the decoder produces samples from the learned distribution. This means the unconditional model is implemented as a decoder-only transformer, while both conditional models use the full encoder--decoder architecture.
Across all models, each transformer component has 4 attention heads, a model dimension of 256 (of which 8 are reserved for a sinusoidal position encoding), 4 feedforward layers of dimension 1024, and a dropout rate of 0. All feedforward activations are ReLU~\cite{DBLP:journals/corr/abs-1803-08375}. Each model is trained for 500 epochs with a batch size of 500 using the Adam optimizer with a learning rate of $1\times 10^{-4}$, scheduled to decay exponentially with $\gamma=0.999$. A schematic overview of the specific model implementation is provided in Figure~\ref{fig:vae_schematic}.

\begin{figure}[t]
  \centering
  \includegraphics[width=0.8\linewidth]{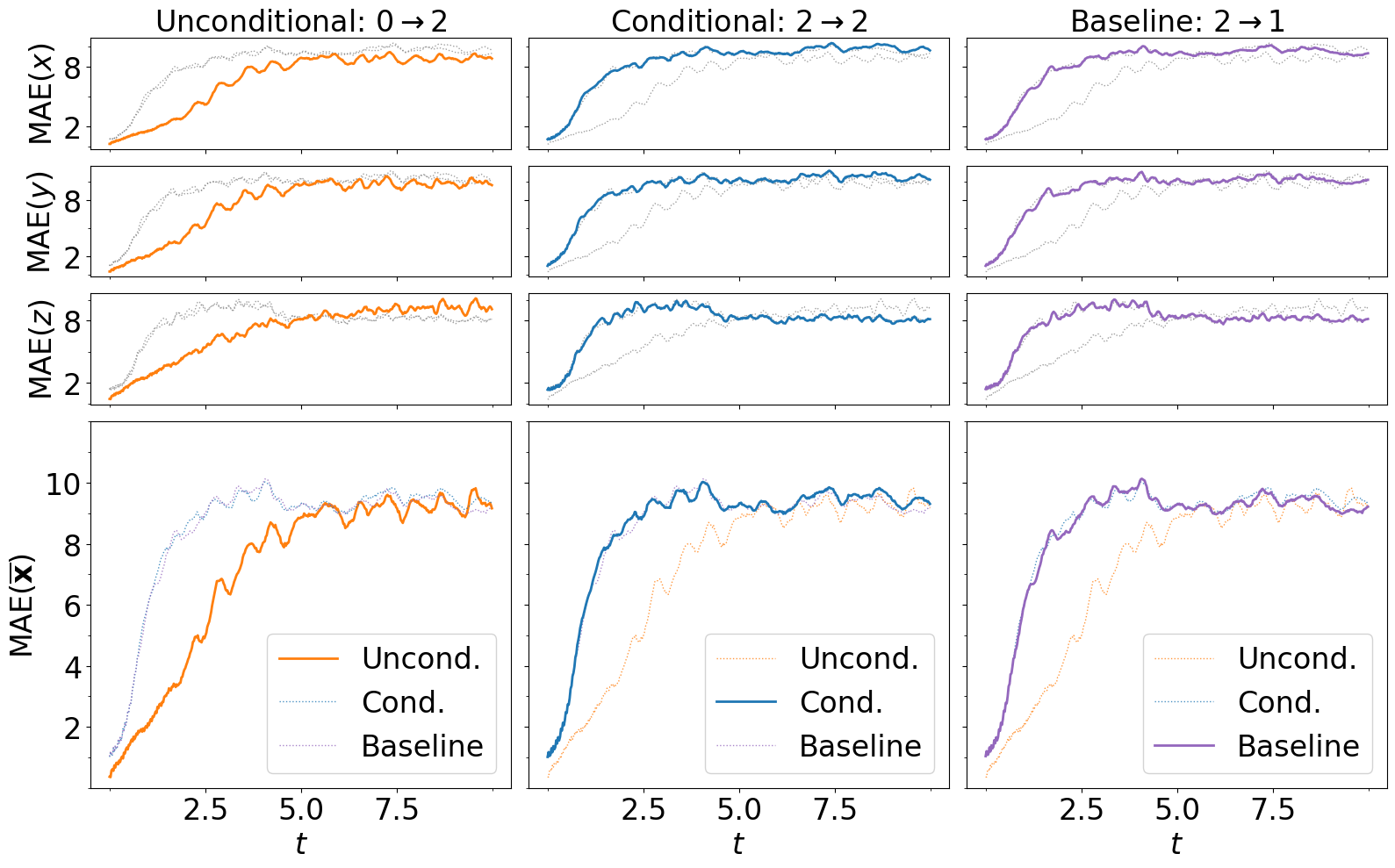}
  \caption{Mean absolute error (MAE) across 500 different initial conditions for Lorenz--63; for each individual component in the first 3 rows, and for the mean across all components below. Each plot redundantly overlays the other columns' MAEs for comparison.}
  \label{fig:l63_mae}
\end{figure}

\subsection{Lorenz--63 model}
\label{subsec:results_l63}
Consider the Lorenz--63 model
\begin{equation}
\begin{aligned}
\frac{dx}{dt} &= \sigma (y - x),\\
\frac{dy}{dt} &= x(\rho - z) - y,\\
\frac{dz}{dt} &= xy - \beta z,
\end{aligned}
\label{eq:lorenz63}
\end{equation}
with $\sigma = 10$, $\rho = 28$, and $\beta = 8/3$. For this choice of parameters, all orbits are attracted to a compact strange attractor with Kaplan--Yorke dimension $2.06$, and exhibit exponential sensitivity to perturbations in the initial conditions. We integrate the system \eqref{eq:lorenz63} using RK4 with fixed time step $\Delta t = 2.5\times10^{-2}$, 
and generate a trajectory of $10^5$ steps. This produces $4\times10^6$ resolved states from which we extract contiguous $n$-step windows to construct the empirical joint distribution $p(x_t,\,x_{t-\Delta t}\,\ldots\,x_{t-(n-1)\Delta t})$ used for training.
A total of $10^6-(n-1)$ joint windows are sampled uniformly in $t=[0,\, 2.5\times(10^4-(n-1)10^{-2})]$ from the attractor for training. For testing, we consider 500 unseen 
initial conditions drawn uniformly in \(t=[2.5\times10^4,\, 10^5]\) forming 
nonoverlapping trajectories. For the distribution adherence test we 
instead use a smaller train--test boundary of $10^3$ and forecast the 
single trajectory segment in $t=[10^3,\, 1.5\times10^3]$.

\begin{figure}[t!]
  \centering
  \includegraphics[width=0.7\linewidth]{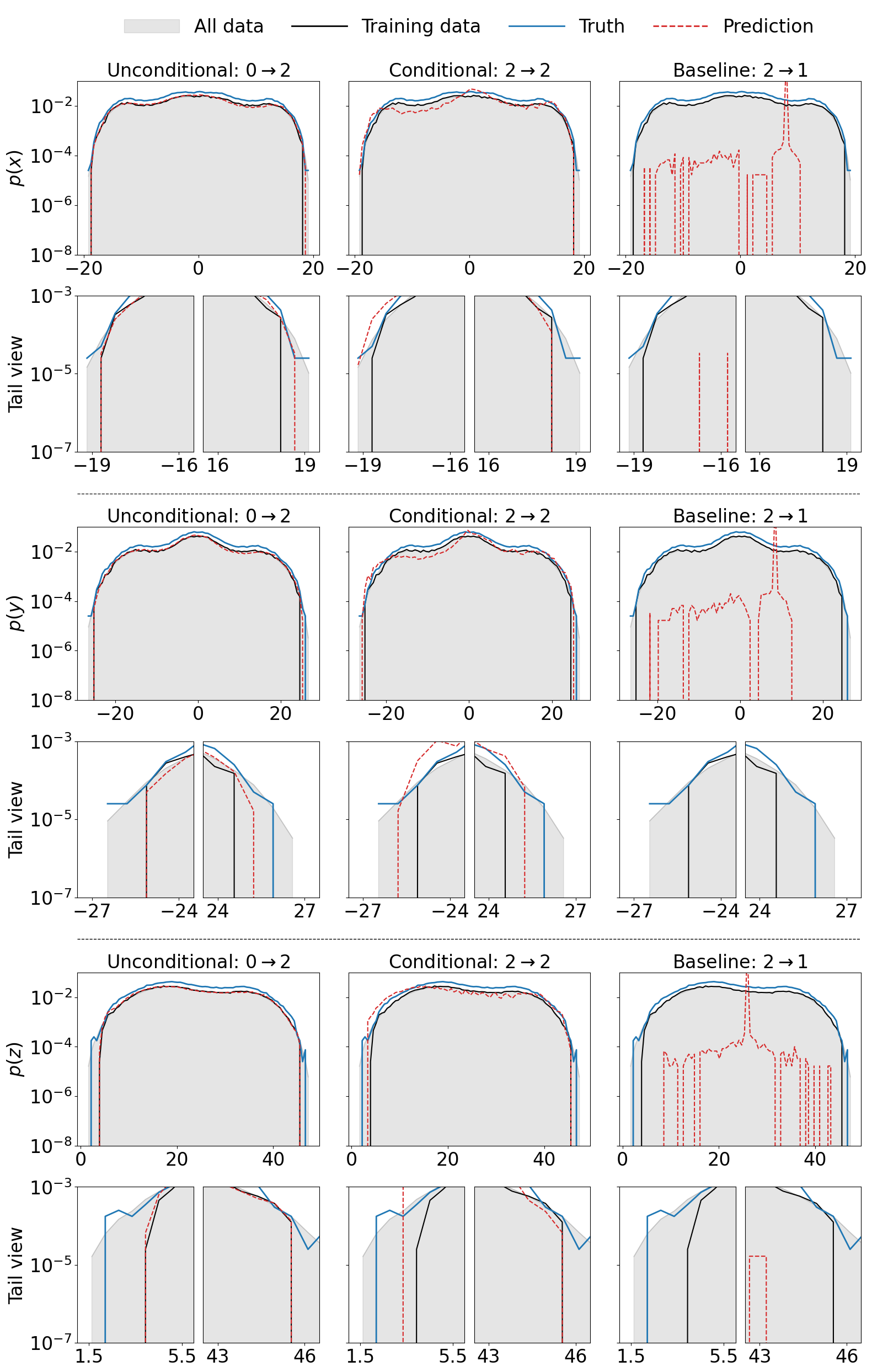}
  \caption{
  Log scale probability density (estimated using relative frequencies) computed over entire synthetic dataset (grey), training set (black), a single trajectory (blue), and the corresponding predicted trajectory (red) for all 3 components of Lorenz--63. Below each component's histogram is a zoomed view for its tails (extreme values).}
  \label{fig:l63_hist}
\end{figure}

In Figures~\ref{fig:l63_single}–\ref{fig:l63_hist} we summarize the results of all three configurations defined in Section \ref{sec:results} on Lorenz--63. 
We first examine single-trajectory behaviour in Figure.~\ref{fig:l63_single}.
All three configurations track the reference trajectory well over the 
first Lyapunov time, preserving the geometry of the attractor in phase space.
By the second Lyapunov time, however, the unconditional joint model maintains its adherence to the reference trajectory, while the conditional joint and baseline models exhibit divergence. Past the fifth Lyapunov time, both joint models continue to reproduce the attractor structure, whereas the baseline model collapses towards the center of a lobe.

For a quantitative assessment we compute the mean absolute error (MAE) across an ensemble of 500 test initial conditions drawn from the attractor but not used for training. Each initial state is forecasted for 10 time units (400 steps at $\Delta t = 2.5\times10^{-2}$). The MAE at each forecast lead time is averaged over trajectories and reported separately for each component and for the componentwise mean in Figure~\ref{fig:l63_mae}.
Here, the unconditional approach dominates both the conditional joint and baseline: initial error is similar initially, but grows at a significantly slower rate over the first Lyapunov time, before converging to the common error level across all configurations by the end of the second. Layering the joint methodology onto the baseline conditional model does not appear to confer any advantage for this system, as observed by the near-identical performance of both the conditional joint and baseline models.

To assess long-term forecasting accuracy and reliability, we integrate a single trajectory for 1500 time units starting from the state immediately following the 1000-time-unit training interval. Each forecasting configuration is run autoregressively over the same horizon, and we compare the resulting distributions of state values.
Figure~\ref{fig:l63_hist} shows log scale histograms for all three components, including: (i) the full synthetic dataset, (ii) the training set, (iii) the long reference trajectory, and (iv) the corresponding predicted trajectories for each model. Tail-focused zooms highlight extreme events.

Both joint models reproduce the bulk distributions well and substantially improve tail behaviour relative to the baseline. The baseline conditional model underrepresents extremes and develops visible distortions in the wings of the attractor.
The conditional joint model achieves the closest match to the reference tails across all components, whereas the unconditional model demonstrates the closest distribution adherence overall. This indicates that the joint generative formulation not only enhances short-term skill but also yields more statistically consistent long-range emulations. Moreover, by more faithfully representing tail probabilities and rare events beyond the typical range explored during training, the joint models suggest a path toward data-driven emulators that can, in principle, better approximate ``grey swan'' events: physically admissible but extremely rare extremes that may be absent from the historical record. This stands in contrast to many state-of-the-art neural network weather models, which have been shown to systematically underestimate such out-of-distribution grey swan tropical cyclones when stronger events are excluded from the training data~\cite{sun2025can}, highlighting the importance of improved tail modelling for robust risk assessment.

\subsection{Kuramoto--Sivashinsky equation}
\label{subsec:results_ks}

To evaluate performance on high-dimensional multiscale chaos, we consider the one-dimensional Kuramoto--Sivashinsky equation
\begin{equation}
\frac{\partial u}{\partial t} +u\frac{\partial u}{\partial x} + \frac{\partial^2 u}{\partial x^2}  + \frac{\partial^4 u}{\partial x^4}  = 0,\quad x\in[-25,25],\quad t>0
\label{eq:ks}
\end{equation}
with initial condition 
\begin{equation}
u(x,0) = \sin(x)\exp\!\left[-\frac{(x-10)^2}{2}\right],
\end{equation}u
and periodic boundary conditions. This setting generates spatiotemporal chaotic dynamics with a broad spectrum of unstable and dissipative modes.  We discretize \eqref{eq:ks} using second-order finite differences on a evenly-spaced spatial grid with spacing $\Delta x = 50/200 =0.25$. We exclude the upper boundary to produce a $199$-dimensional semi-discrete system. Time integration is performed using the two-step Adams--Bashforth method with $\Delta t = 10^{-1}$,  which sufficiently resolves the convective and dissipative scales characteristic of KS turbulence. After an initial transient, the solution reaches a statistical steady state associated with dynamics on a compact strange attractor with Kaplan--Yorke dimension 8.67. From the statistical stationary regime, we extract $10^6-(n-1)$ overlapping $n$-step joint windows for training in $t=[0,\, 10^5-(n-1)10^{-1}]$.
Similarly to the Lorenz--63 case, we evaluate on 500 unseen initial conditions forming nonoverlapping trajectories outside the training window, with the distribution adherence tests using a smaller train--test boundary of $2.5\times10^3$ and forecast endpoint of $6.25\times10^3$.

\begin{figure}[t]
  \centering
  \includegraphics[width=0.8\linewidth]{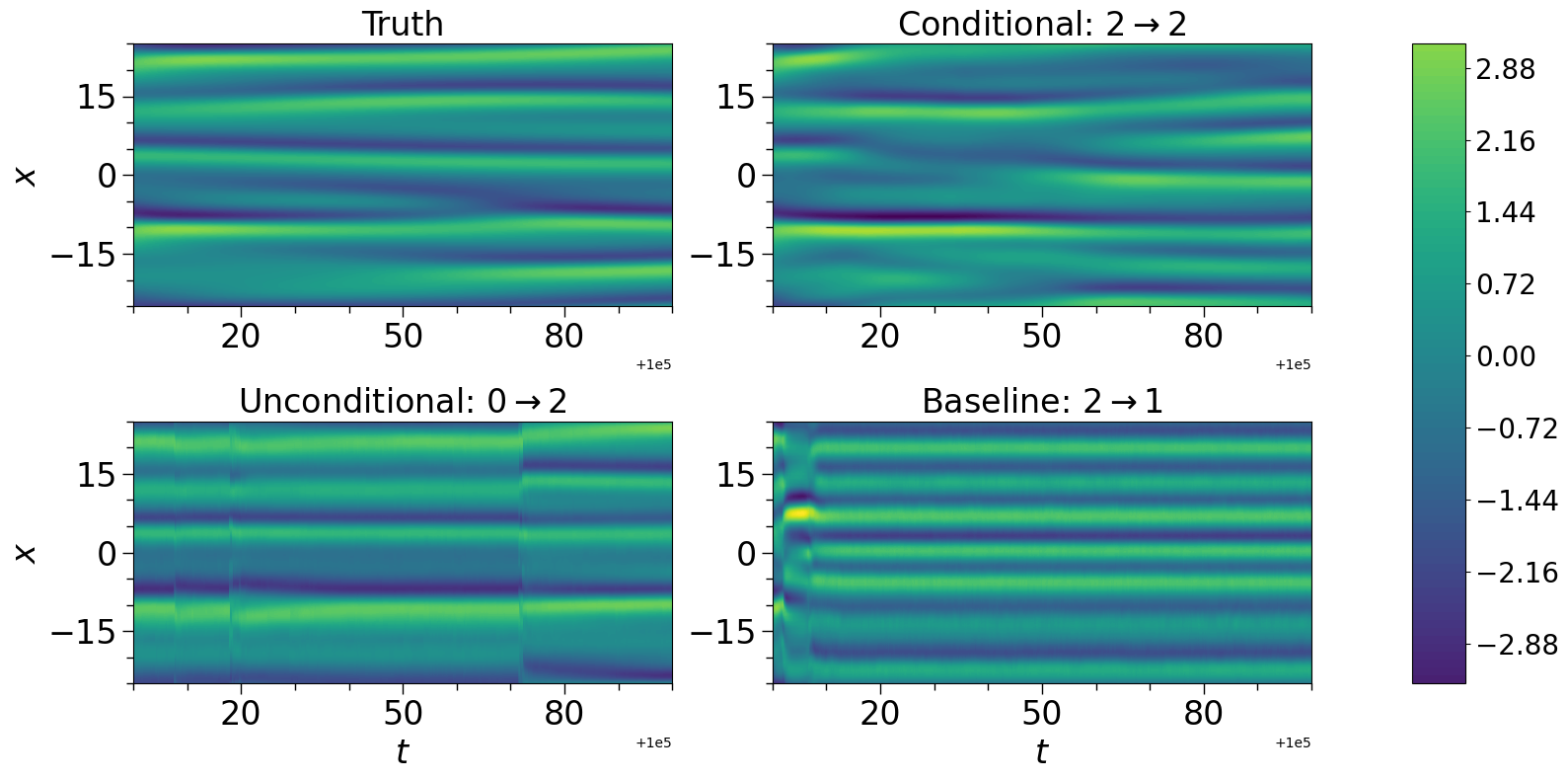}
  \caption{Single trajectory of the Kuramoto--Sivashinsky equation, 100 time units forward from \(t=10^{5}\). Clockwise from top left: reference data, conditional joint model, baseline model, unconditional joint model. Visual jumps in the unconditional output (bottom left) are sharp corrections resulting from prioritizing error minimization over smoothness during inference. This tradeoff can be regulated by adjusting ensemble size in Alg.~\ref{alg:inference} and/or optimization depth in Alg.~\ref{alg:zopt}.}
  \label{fig:ks_single}
\end{figure}

\begin{figure}[t]
  \centering
  \includegraphics[width=0.8\linewidth]{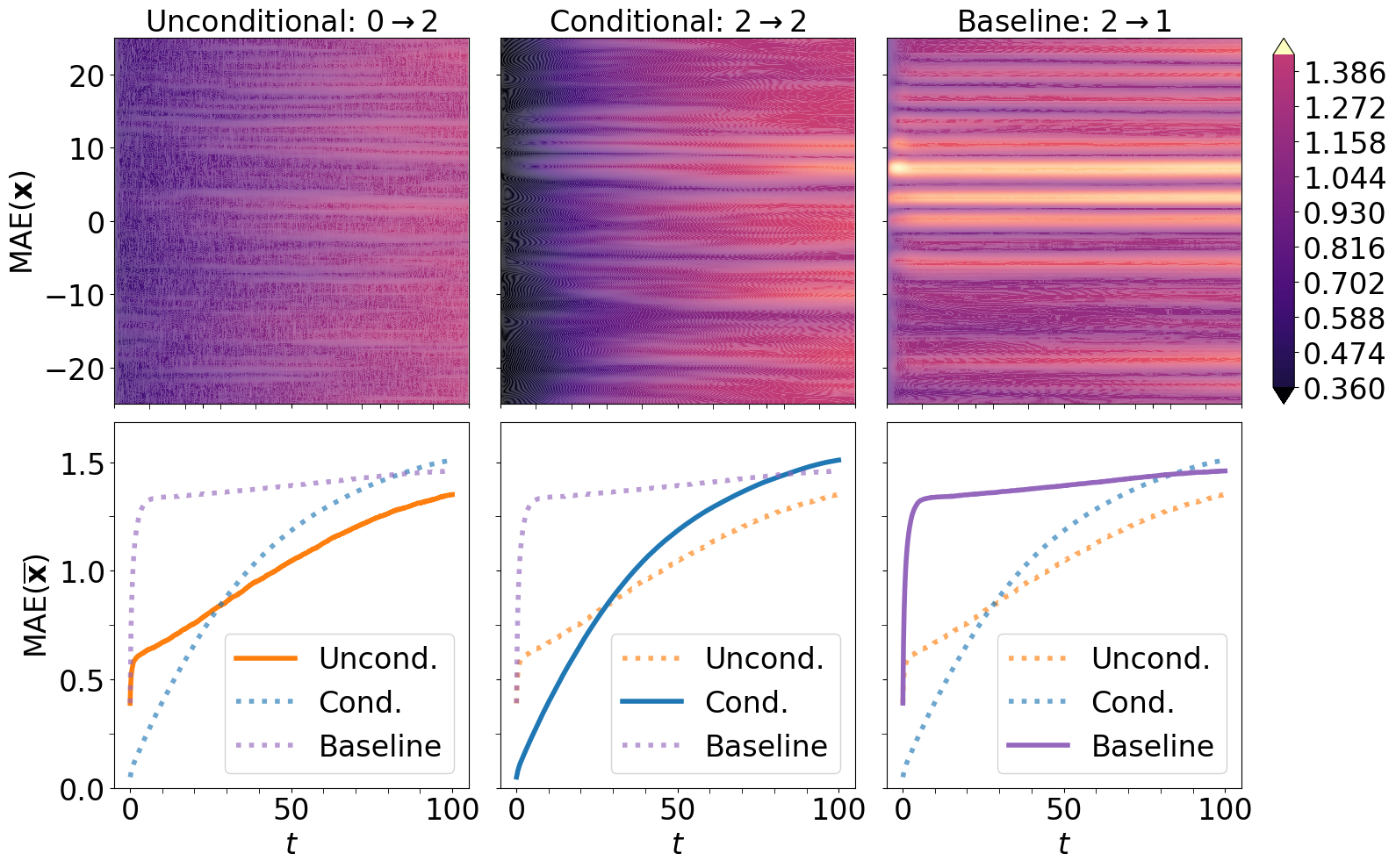}
  \caption{MAE across 500 different initial conditions for KS; for each individual component as a heatmap in the first row, and for the mean across all components below. The componentwise mean MAE plots redundantly overlay the other columns' MAEs for comparison.}
  \label{fig:ks_mae}
\end{figure}

Figure~\ref{fig:ks_single} shows representative spatiotemporal fields over a 1000-step forecast for each configuration. The joint models more faithfully retain the coherent structures and fine-scale patterns of the reference KS field, while the baseline tends to smear small-scale features and exhibits earlier loss of phase information.
For a systematic comparison, we again compute MAE over an ensemble of 500 test initial conditions, each integrated forward for 10 time units (100 steps at $\Delta t = 10^{-1}$). MAE is reported as a function of time for each spatial component (heatmap) and for the spatially averaged error in Figure~\ref{fig:ks_mae}.
Across the forecast horizon, both joint configurations dominate the baseline in MAE: the joint conditional model has the lower initial error and growth in the first Lyapunov time, while the unconditional model stands out in intermediate times where nonlinear interactions are strongest.

\begin{figure}[t]
\centering
  \includegraphics[width=0.8\linewidth]{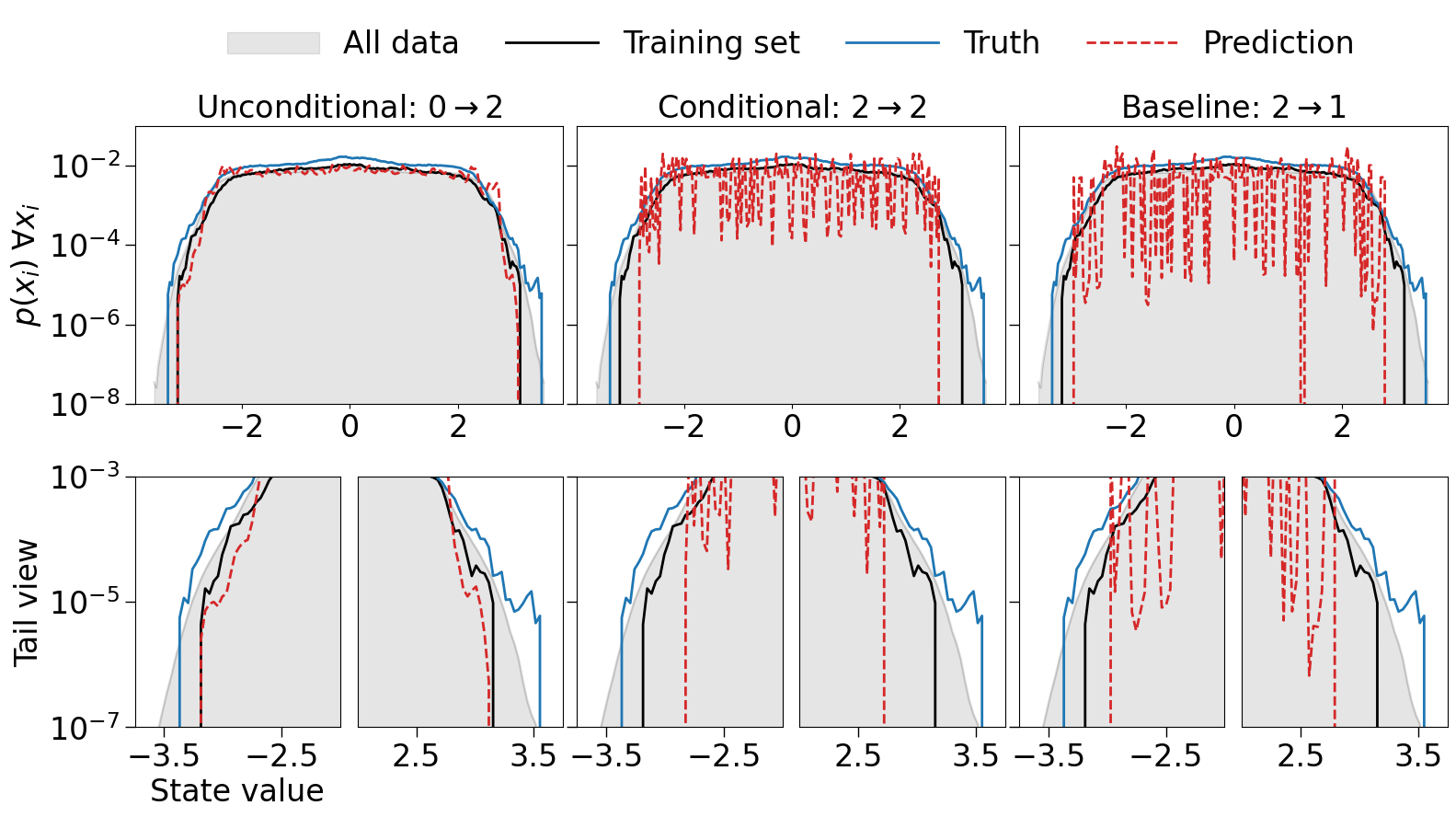}
  \caption{
  Log scale distribution of all values over entire synthetic dataset (grey), training set (black), a single trajectory (blue), and the corresponding predicted trajectory (red) for all values of KS in aggregate. Below each component's histogram is a zoomed view for its tails (extreme values).}
  \label{fig:ks_hist}
\end{figure}

Long-term statistical consistency for KS is assessed in the same manner as for Lorenz--63. We generate a long reference trajectory, 50\% longer than the span of the training range, from the stationary regime and compare it with the corresponding long-horizon forecasts produced by each configuration.
In Figure~\ref{fig:ks_hist} we show log scale histograms of the aggregated KS field values over all spatial locations and times. As before, we include the full synthetic dataset, the training subset, the evaluation trajectory, and the forecasts.

Here, the unconditional model strictly dominates both the conditional joint and baseline models, with the latter two performing similarly.

\subsection{Error prediction metrics}
The proposed joint generative framework provides, in addition to point forecasts, a rich ensemble of joint samples at each time step. Section~\ref{subsec:jgf_uq} introduced three uncertainty quantification (UQ) metrics derived from this ensemble: (i) the ensemble variance $\sigma_{\text{ens}}$ in Eq.~\eqref{eq:ensvar_def}, (ii) the short–horizon autocorrelation $AC$ in Eq.~\eqref{eq:ac_def}, and (iii) the cumulative Wasserstein drift reconstruction $\mathrm{WD}_{\mathrm{recon}}$ in Eq.~\eqref{eq:wdrecon_def}. Here, we evaluate their ability to \emph{predict} pointwise forecast error without access to ground truth at inference time. This inherently gives an a priori estimate of the quality of a forecast model without running extensive evaluations. 

For each trajectory and configuration of interest (we focus solely on the unconditional joint model), we perform linear regressions in which the response variable is the MAE at each time step and the regressors are the UQ metrics. Specifically, we consider four regressions per trajectory: each metric individually, and all three metrics jointly in a multiple regression. The baseline and conditional models are omitted from this analysis because they do not produce a joint ensemble from which these metrics can be computed.
\begin{figure}[t]
\centerline{\hspace{1.3cm}Lorenz--63\hspace{5cm} Kuramoto--Sivashinsky}
 \centerline{
 \includegraphics[width=0.4\linewidth]{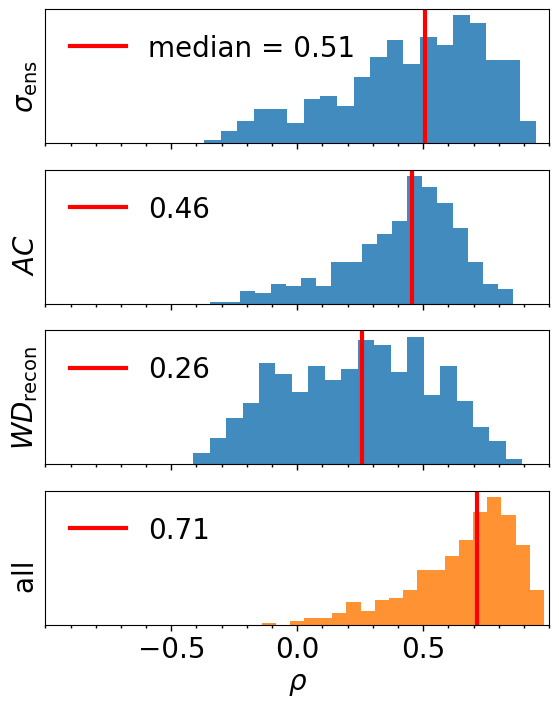}\hspace{1cm}
 \includegraphics[width=0.4\linewidth]{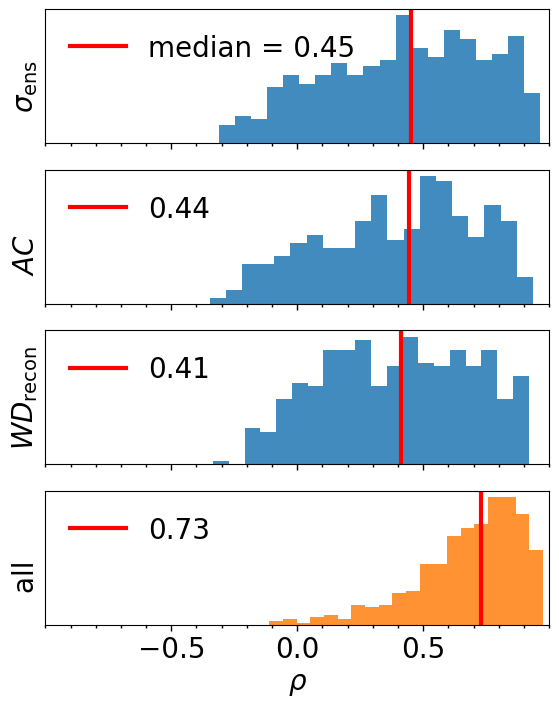}}
 \caption{Histograms of Pearson correlation coefficients \(\rho\) over 500 time series of various linear regression configurations: 3 single-variable (blue), and one multiple regression with all 3 as regressors (orange). Each median \(\rho\) is indicated by the vertical red line. The regressors are defined in Section~\ref{subsec:jgf_uq}: ensemble variance $\sigma_\text{ens}$ \eqref{eq:ensvar_def}, autocorrelation $AC$ \eqref{eq:ac_def}, and Wasserstein distance reconstruction $WD_\text{recon}$ \eqref{eq:wd_def}.}
  \label{fig:mocs}
\end{figure}
\begin{figure}[t]
\centerline{\hspace{1.3cm}Lorenz--63\hspace{5cm} Kuramoto--Sivashinsky}
 \centerline{
 \includegraphics[width=0.4\linewidth]{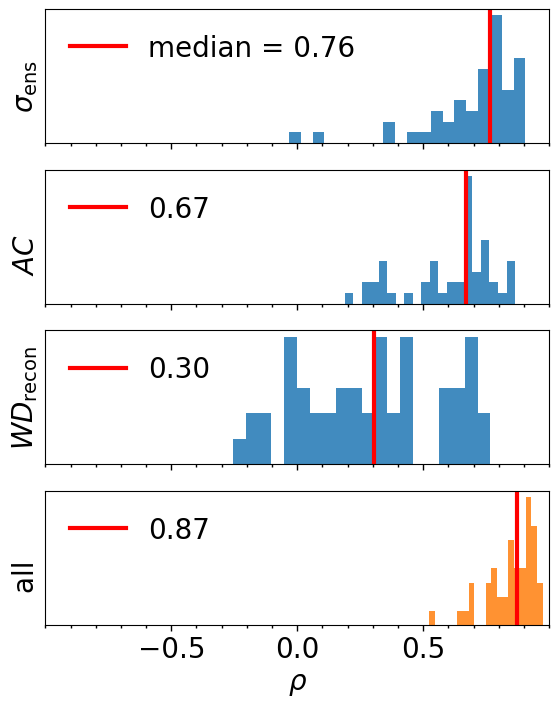}\hspace{1cm}
 \includegraphics[width=0.4\linewidth]{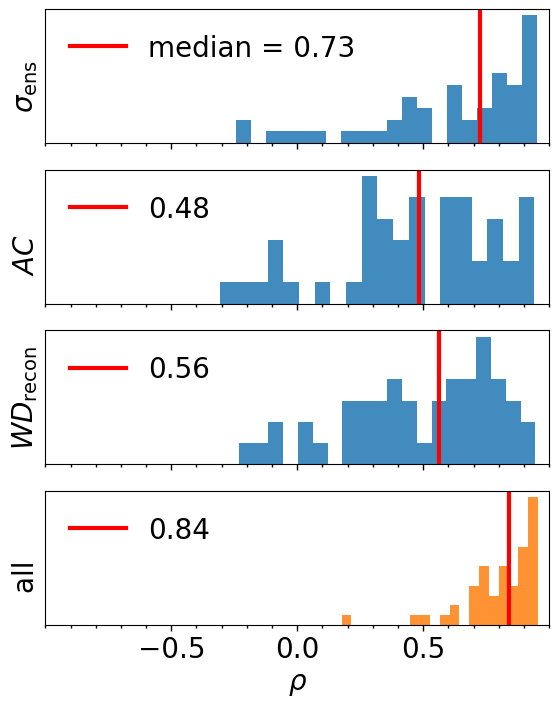}}
 \caption{Histograms of Pearson correlation coefficients \(\rho\) over 50 time series mean from ensembles of size 10 of various linear regression configurations: 3 single-variable (blue), and one multiple regression with all 3 as regressors (orange). Each median \(\rho\) is indicated by the vertical red line. The regressors are defined in Section~\ref{subsec:jgf_uq}: ensemble variance $\sigma_\text{ens}$ \eqref{eq:ensvar_def}, autocorrelation $AC$ \eqref{eq:ac_def}, and Wasserstein distance reconstruction $WD_\text{recon}$ \eqref{eq:wd_def}.}
  \label{fig:coms}
\end{figure}

In Figure~\ref{fig:mocs} we show histograms of cross-validated Pearson correlation coefficient values \(\rho\) across 500 different initial conditions for Lorenz--63 and KS. Any single metric alone yields relatively low explanatory power (median $\rho \leq 0.51$), indicating that no single scalar diagnostic fully captures the structure of the forecast error. When all three metrics are used together, however, the median $\rho$ rises beyond 0.7 for both systems. This demonstrates that ensemble variance, autocorrelation, and Wasserstein drift provide complementary information and, when combined, can explain a substantial fraction of the pointwise error variance.

Finally, in Figure~\ref{fig:coms} we report \(\rho\) values when the regressions are performed on the mean trajectory obtained by averaging 10 independent realizations across 50 unique initial conditions. This results in 50 nonoverlapping trajectories used for 50 instances of each of the four regression. In this aggregated setting, a similar trend if observed with higher correlation values overall, with the multiple regression achieving median \(\rho\ \geq 0.84\) for both systems. These results suggest that the UQ metrics based on joint probability are particularly effective at predicting forecast error in an ensemble setting, strengthening the case for modelling temporal windows jointly rather than conditionally alone.

\section{Conclusion}
\label{sec:conclusion}

This work introduced a fully generative perspective on forecasting chaotic dynamical systems by modelling short temporal windows as joint probability distributions and extracting next-step predictions via marginalization. By reframing forecasting as a generative task rather than a conditional regression problem, the proposed framework enables richer representations of temporal dependencies, enhances short–term predictive skill, and yields markedly improved long-term statistical fidelity for both low–dimensional and high-dimensional chaotic systems. Across Lorenz--63 and the Kuramoto--Sivashinsky equation, the unconditional joint model consistently outperforms a baseline next–step model, particularly in its ability to maintain attractor geometry, reproduce long-term PDFs near the tails, and suppress spurious divergence over long horizons. Furthermore, the joint generative formulation naturally provides intrinsic uncertainty quantification metrics (ensemble variance, autocorrelation, and Wasserstein drift) that collectively predict a substantial fraction of pointwise forecast error without requiring access to ground truth. Together, these results demonstrate that joint generative modelling provides a unified, coherent, and effective approach for probabilistic forecasting in nonlinear dynamical systems that can capture tail statistics, highlighting its potential for modelling rare and extreme events. 

Despite these strengths, several limitations warrant consideration. First, joint modelling introduces additional computational overhead due to the need to sample and compare trajectory segments, which may become burdensome as the state dimension or temporal window length increases. Second, the quality of the marginalization-based inference procedure depends on the density of the sampled joint point cloud; if the point cloud is too sparse, nearest-neighbour matching may fail to provide meaningful conditioning. Third, while long-term statistics are well reproduced, the framework remains data-driven and ultimately inherits any biases or coverage limitations present in the training distribution. As with most generative surrogates, performance may degrade under strong distribution shift or in regimes poorly represented in the training data. Addressing these limitations will be essential for deploying joint generative forecasting in real-world scientific settings.
A key challenge revealed by the high-dimensional KS experiments is scalability. Although the unconditional joint model performs well even in 199 dimensions, the curse of dimensionality manifests itself during the inference stage: large point clouds are required for reliable matching, and sampling cost rises accordingly. Latent optimal control provides partial relief by replacing combinatorial search with continuous optimization in latent space, but further architectural and algorithmic innovations are needed. Efficient sequence-aware generative models, structured latent spaces, and dimension-reduced temporal embeddings may offer pathways to reduce sampling requirements. Moreover, because the joint distribution factorizes across the temporal vs.\ spatial axes, exploiting tensorized \cite{PhysRevE.110.015310,DEKTOR2023112378,adaptive_rank,DEKTOR2021110295} or factored parameterizations could allow scaling to domains with thousands to millions of degrees of freedom, such as climate, ocean, or turbulence simulations.

The proposed generative forecasting framework can be extended in several directions. First, physical constraints, such as invariances, conservation laws, or energy-based regularization terms, can be incorporated to improve extrapolation and stability in long-horizon forecasts \cite{Raissi1,GK2021}. Second, integrating adaptive temporal window lengths could allow the model to dynamically adjust the memory of the joint distribution based on local flow regimes. Third, the demonstrated ability of the joint model to reproduce extreme-event tail statistics suggests promise for grey swan prediction and risk assessment in systems where rare events play an outsized role. This aligns with recent findings that standard neural forecasting models systematically underpredict out-of-distribution extremes, such as grey swan tropical cyclones. Finally, combining the joint generative framework with hybrid physical--neural architectures, flow-matching techniques, or diffusion-based priors may further enhance both scalability and uncertainty quantification. Overall, the results presented here position joint generative forecasting as a principled and extensible foundation for next-generation data-driven modelling of complex dynamical systems.

\section*{Acknowledgements}
\noindent
This work was supported by the U.S. Department of Energy (DOE) under grant ``Resolution-invariant deep learning for accelerated propagation of epistemic and aleatory uncertainty in multi-scale energy storage systems, and beyond,'' contract number DE-SC0024563.

\bibliographystyle{plain}
\bibliography{main}

\end{document}